\documentclass{article}
\usepackage{spconf,amsmath,epsfig}

\let\OLDthebibliography\thebibliography
\renewcommand\thebibliography[1]{
  \OLDthebibliography{#1}
  \setlength{\parskip}{0pt}
  \setlength{\itemsep}{0pt plus 0.3ex}
}
\usepackage{cite}
\usepackage{amsmath,amssymb,amsfonts}
\usepackage{algorithmic}
\usepackage{graphicx}
\usepackage{textcomp}
\usepackage{xcolor}
\usepackage{algorithm}
\usepackage{algorithmic}
\usepackage{subfigure}

\begin{document}\sloppy

\def\x{{\mathbf x}}
\def\L{{\cal L}}

\title{Deep Unsupervised Hashing by Distilled Smooth Guidance}
\name{Xiao Luo$^{1,2*}$\thanks{The work was done when Xiao Luo and Daqing Wu interned in Damo Academy, Alibaba Group. This work was supported by The National Key Research and Development Program of China (No.2016YFA0502303) and the National Natural Science Foundation of China (No.31871342). $^*$ Equal contribution. $^\dagger$ Corresponding authors:
Chong Chen~(cheung.cc@alibaba-inc.com) and Minghua Deng~(dengmh@pku.edu.cn)}, Zeyu Ma$^{3*}$, Daqing Wu$^{1,2*}$, Huasong Zhong$^2$, Chong Chen$^{1,2}$, Jinwen Ma$^1$, Minghua Deng$^{1}$}

\address{$^1$School of Mathematical Sciences, Peking University, China \\$^2$Damo Academy, Alibaba Group, Hangzhou, China \\$^3$School of Computer Science and Technology, Harbin Institute of Technology, Shenzhen, China}

\maketitle

\begin{abstract}
Hashing has been widely used in approximate nearest neighbor search for its storage and computational efficiency. Deep supervised hashing methods are not widely used because of the lack of labeled data, especially when the domain is transferred. Meanwhile, unsupervised deep hashing models can hardly achieve satisfactory performance due to the lack of reliable similarity signals. To tackle this problem, we propose a novel deep unsupervised hashing method, namely Distilled Smooth Guidance (DSG), which can learn a distilled dataset consisting of similarity signals as well as smooth confidence signals. To be specific, we obtain the similarity confidence weights based on the initial noisy similarity signals learned from local structures and construct a priority loss function for smooth similarity-preserving learning. Besides, global information based on clustering is utilized to distill the image pairs by removing contradictory similarity signals. Extensive experiments on three widely used benchmark datasets show that the proposed DSG consistently outperforms the state-of-the-art search methods.
\end{abstract}


\section{Introduction}
Deep learning-based hashing methods can be divided into supervised hashing and unsupervised hashing\cite{luo2020survey}. At the early stage, many researchers mainly focused on the supervised hashing methods, which utilize semantic labels to greatly improve the performance of image retrieval\cite{xia2014supervised}. However, supervised hashing methods are difficult to be applied in practice when there is not enough labeled data, especially when the domain is transferred. To solve this problem, several deep learning-based unsupervised methods were proposed, including deep binary descriptors (DeepBit)\cite{lin2016learning}, semantic structure-based unsupervised deep hashing (SSDH)\cite{yang2018semantic} and unsupervised deep hashing by distilling data pairs (DistillHash)\cite{yang2019distillhash}. 

Although deep learning-based unsupervised hashing methods can be applied on unlabelled data, they still have evident limitations. \cite{hu2017pseudo} takes clustering information to generate pairwise pseudo-labels. SSDH further studies the deep feature statistics empirically from a pre-trained model and captures the semantic relationships across different data points. Specifically, it selects image pairs with confident pseudo-labels to guide the training of the model. Nevertheless, SSDH discards most image pairs which are hard to be decided whether they are semantically similar or dissimilar, which causes much information loss and thus limits the performance of the model in further image retrieval. In general, these methods only consider either the local information of similarity signals or the global information such as clustering labels but do not consider this task comprehensively. 


To tackle the above issues, we propose a novel method, which comprehensively explores correlations among image pairs. First of all, for the correlation among different samples, we adopt the pre-trained deep convolutional neural network (CNN) to generate features for the input images. Then we compute the pairwise cosine distance and construct the similarity pseudo graph. We take all the image pairs into consideration. To alleviate the effect of wrong pseudo labels, different weights are assigned to image pairs according to the confidence of the pseudo labels. Furthermore, for the global robustness, we adopt clustering on the deep image features and then obtain another similarity graph. According to these two similarity graphs, image pairs with different correlation identification are considered contradictory and thus distilled. Finally, we design a deep neural network based on the distilled data pair set and adopt a deep learning framework to perform the deep representation and hash code learning simultaneously. Our main contributions can be summarized as follows:
\begin{itemize}
	\item We introduce two similarity graphs based on the local (i.e., pairwise cosine similarity) and global information (i.e., clustering) respectively and then obtain the ensemble similarity graph by distilling the contradictory pairs. 
	\item We construct a priority loss function for similarity-preserving learning, which prioritizes confident image pairs over fuzzy image pairs based on their pairwise distance to learn deep representations smoothly.
	\item Experiments on three popular benchmark datasets show that our method DSG outperforms current state-of-the-art unsupervised hashing methods by a large margin.
\end{itemize}

\begin{figure*}
\centering
\includegraphics[width=16cm,keepaspectratio=true]{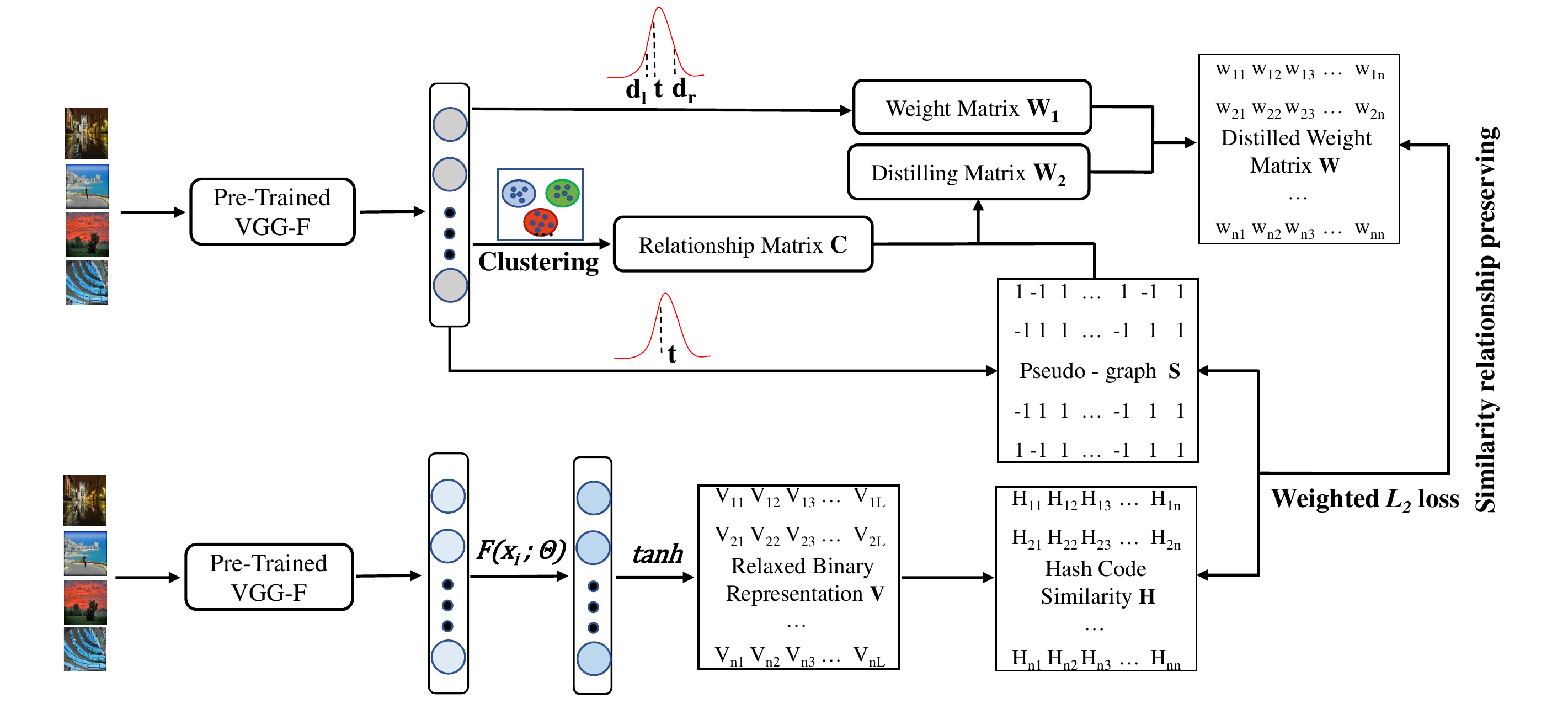}
\caption{The framework of our model. First, deep features of the dataset are extracted through the pre-trained VGG-F network. Second, the pseudo-graph $S$ and the smooth weight matrix $W_1$ are constructed based on the local information of the cosine distance distribution of deep features. Third, the relationship matrix $C$ is constructed based on the global information of the K-means clustering on deep features and the distilling matrix $W_2$ is obtained by comparing the relationship matrix $C$ with the pseudo-graph $S$. Lastly, iterations are done aiming to minimize the weighted $L_2$ loss for preserving similarity relationship of image pairs (i.e., Equation \ref{eq5}) by using the mini-batch stochastic gradient descent method.}
\label{fig:framwork}
\end{figure*}
\section{Related Work}
{\bf{Deep Supervised Hashing.}} Deep supervised hashing methods usually map the data points into Hamming space where the semantic similarities can be preserved by learning a deep neural network\cite{luo2020survey}. Typically, as the first supervised deep hashing method, CNNH\cite{xia2014supervised} splits the hash learning into two stages based on a convolutional neural network. Hash codes are learned on the first stage, a specific deep network is learned on the second stage to map the input samples to the learned hash codes. Deep Supervised Hashing\cite{liu2016deep} uses a loss function with product form based on the pairwise similarities and Hamming distances, which can be trained by the end-to-end back propagation algorithm.

\noindent {\bf{Deep Unsupervised Hashing.}}
Unsupervised deep hashing methods aim to turn unsupervised problems into supervised problems by constructing pseudo labels based on deep features. Semantic Structure-based Unsupervised Deep Hashing (SSDH)\cite{yang2018semantic} uses a specific truncated function on the pairwise distances and constructs the similarity matrices. DistillHash\cite{yang2019distillhash} improves the performance of SSDH by distilling the data pairs for confident similarity signals. MLS$^3$RDUH \cite{ijcai2020-479} utilizes the intrinsic manifold structure in feature space to reconstruct the local semantic similarity structure, and achieves the state-of-the-art performance.

\section{Method}
In the problem of deep unsupervised hashing, $\mathcal{X}=\{x_i\}_{i=1}^N$ denotes the training set with $N$ samples without label annotations, it aims to learn a hash function $$\mathcal{H}: x\rightarrow b \in\{-1,1\}^L,$$
where $x$ is the input sample and $b$ is a compact $L$-bit hash code. 
It is noticed that $x_i\in \mathbb{R}^d$ is the normalized extracted deep feature for the $i$-th image through the pre-trained neural network by removing the last fully-connected layer. Here we use VGG-F\cite{simonyan2014very} to be consistent with other articles. 

\subsection{Pseduo-Similarity Graph}

In our model, the pseudo-similarity graph is constructed firstly. The pseudo-similarity graph is used to capture pairwise similarity information from a local perspective. Based on the pre-trained deep feature $x_i $, the cosine distance between the i-th and the j-th samples can be computed by $d_{ij}= 1 - \frac{x_i\cdot x_j}{||x_i||_2||x_j||_2}$. We set a large threshold $t$, and consider data points with the cosine distance smaller than $t$ as potential similar and data points with the cosine distance larger than $t$ as potential dissimilar. Based on the threshold $t$, we construct the pseudo-graph $S$ as:

\begin{equation}\label{eq1}
    S_{ij}=\left\{ 
\begin{array}{ll}
 1 & d_{ij}\leq t,  \\ 
 -1 & d_{ij}> t
\end{array}\right.
\end{equation}

Where $S_{ij}$ is set to 1 if points $x_i$ and $x_j$ are potential similar, and -1 if points $x_i$ and $x_j$ are potential dissimilar.

\subsection{Smooth Weight Matrix}
Although pseudo-similarity graph is constructed, the semantic confidence of pseudo-label for each pair is different. In this section, we construct the weight matrix for the pseudo-graph based on the semantic confidence.  

By observing the distribution of cosine distance for deep feature pairs, \cite{yang2018semantic} finds that each distance histogram is similar to two half Gaussian distributions, where $m_l$ and $\sigma_l$ denote the mean and the standard deviation of the first (left half) distribution and $m_r$ and $\sigma_r$ denote the mean and the standard deviation of the second (right half) distribution. Accordingly, we obtain two distance thresholds $d_l= m_l-\alpha \sigma_l $ and $d_r = m_r +\beta \sigma_r$, where the hyper-parameters $\alpha$ and $\beta$ control the values of the distance thresholds $d_l$ and $d_r$ respectively and dictate the percentage of similar points and dissimilar points from all data points respectively as well. According to the theory of confidence interval, the pairs with distances smaller than $d_l$ or larger than $d_r$ have confident semantic similarity information. As a result, we set the weights for confident pairs to 1. 
Since the pairs with distance in the interval $[d_l,d_r]$ probably have no certain semantic information, we obtain the weight as:
\begin{equation}\label{eq:2}
    W_{1,ij}=\left\{ 
\begin{array}{ll}
 \frac{(t - d_{ij})^{2}}{(t - d_l)^{2}} & d_l < d_{ij}\leq t,  \\ 
 \frac{(d_{ij} - t)^{2}}{(d_r - t)^{2}} & t < d_{ij} < d_r, \\
 1 & d_{ij}\leq d_l \ or \ d_{ij}\geq d_r
\end{array}\right.
\end{equation}

From the equation, the weight for each image pair with the distance out of the interval is set to $1$ and the weight for each image pair with the distance in the interval is smaller if their distance is closer to the threshold in a quadratic form. In this way, all the image pairs are taken into consideration with the guidance of the smooth weight matrix. Accordingly, the confidence of their similarity relationship is guided by the smooth weight $W_1$.

\subsection{Pair Distilling based on clustering}
As we know, the obtained pseudo-label is very coarse. In this section, we try to use the clustering method to distill the pairs by removing contradictory results. To be specific, we first use the extracted features$\{x_i\}_{i=1}^N$ to construct a K-means clustering model. Based on the K-means clustering result, we construct a relationship matrix $C$:
\begin{equation}\label{eq3}
   C_{ij}=\left\{ 
\begin{array}{ll}
 1 & c_i = c_j,  \\ 
 -1 & c_i \neq c_j
\end{array}\right. 
\end{equation}

where $c_i$ represents the label of the cluster that the data point $x_i$ belongs to. $c_i \in \{0, 1, ..., K\}$ and $K$ is the parameter that represents the number of clusters. By comparing the relationship matrix $C$ with the pseudo-graph $S$, we obtain the distilling matrix $W_2$ as:
\begin{equation}\label{eq4}
    W_{2,ij} = \left\{
\begin{array}{ll}
 1 & S_{ij} = C_{ij},  \\ 
 0 & S_{ij} \neq C_{ij}
\end{array}\right.
\end{equation}

in which $S_{ij} = C_{ij}$ means that the result of pseudo-graph and clustering for the pair $x_i$ and $x_j$ is consistent. 

After setting the weight of contradictory pairs to zero, the final distilled weight matrix is obtained as $$W=W_1\odot W_2, $$ in which $\odot $ denotes the element-wise product of vectors. 

\subsection{Hash code learning}
For the purpose of preserving the similarity relationship of data points, similar data points are expected to be mapped into similar hash codes and dissimilar data points are expected to be mapped into dissimilar hash codes. The similarity output $H_{N\times N}$ of hash codes is formulated as 
\begin{equation}
H_{i j}=\frac{1}{L} \boldsymbol{b}_{i} \top \boldsymbol{b}_{j}, \quad \boldsymbol{b}_{i} \in\{-1,+1\}^{L}
\end{equation} 
To preserve the obtained semantic structures, we minimize the weighted $L_2$ loss between the hash code similarity and the pseudo-graph. In formulation,
$$\mathcal{L}=\frac{1}{N^2}\sum_{i=1}^N\sum_{j=1}^NW_{ij}(H_{ij}-S_{ij})^2$$
in which $b_i = sign(F(x_i;\Theta))$ and $F(x_i;\Theta)$ denotes the output of the neural network, and $\Theta$ is the parameters. In this way, we can integrate the above loss function into the deep architecture. However, it is infeasible to train the neural network with binary outputs by the standard Back Propagation algorithm because of the ill-posed gradient problem. As a result, $tanh(\cdot)$ is utilized to relieve the binary constraint. 
\begin{algorithm}
\caption{Training algorithm for our model}
\label{alg1}
\begin{algorithmic}[1]
\REQUIRE Training images $\mathcal{I}=\{I_{1}, \dots, I_{N}\}$;\\
\quad \ \ \ Number of clusters: $K$;\\
\quad \ \ \ Cosine distance threshold: $t$
\ENSURE Parameters $\Theta$ for the neural network;\\
\qquad\ Hash codes $B$ for training images.
\STATE Get deep features of $\mathcal{I}$ through VGG-F: $\mathcal{X} = \{x_1,\dots, x_N\}$.
\STATE Construct the pseudo-graph $S$ by Equation \ref{eq1}
\STATE Construct the weight matrix $W_1$ by Equation \ref{eq:2}
\STATE Cluster $\mathcal{X}$ into $K$ different groups by $K$-means. and construct the distilling matrix $W_2$ by Equation \ref{eq3}
\STATE Calculate distilled weight matrix $W$ by Equation \ref{eq4}
\REPEAT
\STATE Sample N points from $\mathcal{X}$ and then construct a mini-batch.
\STATE Calculate the outputs by forward-propagating through the network.
\STATE Update parameters of the VGG-F network by Back propagation by Equation \ref{eq5}
\UNTIL convergence
\end{algorithmic}
\end{algorithm}
Thus we adopt the following objective function:
\begin{equation}\label{eq5}
\begin{array}{c}\min _{\Theta} \mathcal{L}(\Theta)=\frac{1}{m^{2}} \sum_{i=1}^{m} \sum_{j=1}^{m}W_{ij}\left(H_{i j}-S_{i j}\right)^{2} \\ \\ \text {s.t.} \quad H_{i j}=\frac{1}{L} \boldsymbol{v}_{i} \top \boldsymbol{v}_{j}, \quad \boldsymbol{v}_{i}=\tanh \left(F\left(\boldsymbol{x}_{i} ; \Theta\right)\right)\end{array}
\end{equation}
in which $\boldsymbol{v}_i$ denotes the relaxed binary representation. 

For the point $q_i$ not in the training set, its hash code $\boldsymbol{b}_i$ is obtained by directly forward propagating it through the learned neural network.
\begin{equation}
\boldsymbol{b}_{i}=\operatorname{sign}\left(F\left(\boldsymbol{q}_{i} ; \Theta\right)\right)
\end{equation}

\subsection{Optimization}
To optimize the problem, we construct the pseudo-graph $S$ from the pre-trained neural network by using Equation \ref{eq1}. Then the smooth weight matrix $W_1$ and the distilling matrix $W_2$ are constructed to get the distilled weight matrix $W$. Lastly, we minimize Equation \ref{eq5} by using the standard stochastic gradient descent (SGD) method. The whole learning procedure is summarized in Algorithm 1.

\section{Experiments}
	   
	
\begin{table*}
\small
\centering
\caption{MAP for different methods on FLICKR25K, CIFAR-10 and NUS-WIDE datasets.\label{tab:1}}
	\begin{tabular*}{1\textwidth}{@{\extracolsep{\fill}}l|cccc|cccc|cccc@{}}
		\hline
		& \multicolumn{4}{c|}{FLICKR25K} & \multicolumn{4}{c|}{CIFAR-10} & \multicolumn{4}{c}{NUS-WIDE} \\
		\hline
		Method & 16bits & 32bits & 64bits &128bits & 16bits & 32bits & 64bits &128bits & 16bits & 32bits & 64bits &128bits\\
		\hline
		SH & 0.6091 & 0.6105 & 0.6033 & 0.6014 & 0.1605 & 0.1583 & 0.1509 & 0.1538 & 0.4350 & 0.4129 & 0.4062 & 0.4100\\
		SpH & 0.6119 & 0.6315 & 0.6381 & 0.6451 & 0.1439 & 0.1665 & 0.1783 & 0.1840 & 0.4458 & 0.4537  & 0.4926 & 0.5000\\
		SGH & 0.6362 & 0.6283 & 0.6253 & 0.6206 & 0.1795 & 0.1827 & 0.1889 & 0.1904 & 0.4994 & 0.4869 & 0.4851 & 0.4945\\
		DeepBit &0.5934 & 0.5933 & 0.6199 & 0.6349 & 0.2204 & 0.2410 & 0.2521 & 0.2530 & 0.3844 & 0.4341 & 0.4461 & 0.4917\\
		SSDH & 0.7240 & 0.7276 & 0.7377 & 0.7343 & 0.2568 & 0.2560 & 0.2587 & 0.2601 & 0.6374 & 0.6768 & 0.6829 & 0.6831\\
		DistillHash & 0.6964 & 0.7056 & 0.7075 & 0.6995 & 0.2844& 0.2853 &0.2867 & 0.2895 & 0.6667&0.6752& 0.6769 & 0.6747\\
		CUDH & 0.7332 & 0.7426 & 0.7549 & 0.7561 & 0.2856&0.2903 &0.3025&0.3000&0.6996&0.7222&0.7451&0.7418\\
		MLS$^3$RDUH &0.7587 &0.7754 &0.7870 &0.7927 &0.2876 &0.2962 & 0.3139 &0.3117 &0.7056 &0.7384 &0.7629 &0.7818 \\
		
		DSG  &\textbf{0.7994} & \textbf{0.8172} & \textbf{0.8197} & \textbf{0.8245} & \textbf{0.3225} & \textbf{0.3241} & \textbf{0.3453} & \textbf{0.3450} & \textbf{0.7795} & \textbf{0.7981} & \textbf{0.8098} & \textbf{0.8187} \\
		\hline
	\end{tabular*}
\end{table*}

\begin{table*}
\small
\centering
\caption{Ablation analysis on datasets FLICKR25K, CIFAR-10 and NUS-WIDE.\label{tab:2}}
		\begin{tabular*}{1\textwidth}{@{\extracolsep{\fill}}l|cccc|cccc|cccc@{}}
		\hline
		& \multicolumn{4}{c|}{FLICKR25K} & \multicolumn{4}{c|}{CIFAR-10} & \multicolumn{4}{c}{NUS-WIDE} \\
		\hline
		Method & 16bits & 32bits & 64bits &128bits & 16bits & 32bits & 64bits &128bits & 16bits & 32bits & 64bits &128bits\\
		\hline
		DSG-v2  & 0.6896 & 0.6570 & 0.6228 & 0.6178 & 0.2356 & 0.2099 & 0.1704 & 0.1567 & 0.6967 & 0.6736 & 0.6257 & 0.6062\\
		DSG-v1 & 0.7584 & 0.7432 & 0.7355& 0.7240 & 0.2612 & 0.2496 & 0.2441 & 0.1988 & 0.7632 & 0.7718 & 0.7719 & 0.7703\\
		DSG &\textbf{0.7994} & \textbf{0.8172} & \textbf{0.8197} & \textbf{0.8245} & \textbf{0.3225} & \textbf{0.3241} & \textbf{0.3453} & \textbf{0.3450} & \textbf{0.7795} & \textbf{0.7981} & \textbf{0.8098} & \textbf{0.8187} \\
		\hline
	\end{tabular*}
\end{table*}

We implement extensive experiments on three datasets to evaluate our DSG by comparing with several state-of-the-art unsupervised hashing methods.
\subsection{Datasets and Baselines}
CIFAR-10\cite{krizhevsky2009learning} is a dataset for image classification and retrieval, containing 60K images from 10 different categories. For each class, we randomly select 1,000 images as queries and 500 as training images, resulting in a query set containing 10,000 images and a training set made up of 5,000 images. All images except for the query set are used as the retrieval set.
\noindent NUS-WIDE\cite{chua2009nus} contains 269,648 images, each of the images is annotated with multiple labels referring to 81 concepts. The subset containing the 10 most popular concepts is used here. We randomly select 5,000 images as a test set; the remaining images are used as a retrieval set, and 5000 images are randomly selected from the retrieval set as the training set. 
\noindent FLICKR25K\cite{huiskes2008mir} contains 25,000 images collected from the Flickr website. Each image is manually annotated with at least one of the 24 unique labels provided. We randomly select 2,000 images as a test set; the remaining images are used as a retrieval set, from which we randomly select 10,000 images as a training set.

Our method is compared with both traditional hashing methods and state-of-the-art unsupervised deep learning methods. Traditional methods includes SpH\cite{heo2012spherical}, DSH\cite{jin2013density} and SGH\cite{dai2017stochastic}. Deep unsupervised hashing methods includes DeepBits\cite{lin2016learning}, SSDH\cite{yang2018semantic}, DistillHash \cite{yang2019distillhash}, CUDH \cite{2019Clustering}, and MLS$^3$RUDH\cite{ijcai2020-479}. 

\subsection{Implementation Details}
The framework is implemented by Pytorch V1.4. The mini-batch size is set to 24 and the momentum to 0.9. The learning rate is fixed at 0.001. The initial weights of the first seven layers of the neural network are from the model pre-trained with ImageNet, and the last fully-connected layer is learnt from scratch. The parameter $\alpha$ and $\beta$ are all set following \cite{yang2018semantic} and the threshold $t$ to $0.1$. The number of clusters is $70$.

\subsection{Evaluation}
The ground-truth similarity information for evaluation is constructed from the ground-truth image labels: two data points are considered similar if they share the same label (for CIFAR-10) or share at least one common label (for FLICKR25K and NUSWIDE).

The retrieval quality are evaluated by the following three evaluation metrics: Mean Average Precision (MAP), Precision-Recall curve and Top N precision curve.

MAP is a widely-used criteria to evaluate retrieval accuracy. Given a query and a list of $R$ ranked retrieval results, the average precision (AP) for the given query can be computed. MAP is defined as the average of APs for all queries. For datasets FLICKR25K and NUSWIDE, we set $R$ as $5000$ for the experiments. For CIFAR-10, $R$ is set as $50000$. Precision-recall curve reveals the precision at different recall levels and is a good indicator of overall performance. In addition, Top N precision curve, which is the precision curve with respect to the top $K$ retrieved instances, also visualizes the performance from a different perspective . 
\subsection{Overall Performance and Ablation Study}

Table \ref{tab:1} shows the MAPs for different methods on three datasets with hash code lengths varying from 16 to 128. It can be seen that the performances of deep learning-based algorithms are overall better than traditional methods. For our proposed method, we find that DSG has a significant improvement over the state-of-the-art deep learning-based methods in all cases, which implies the superiority of our model. By comparing with MLS$^3$RUDH, which has the best performance of MAP among the deep hashing methods, DSG improves the MAP by 3.18\%, 3.33\% and 3.69\% for 128 bit length on datasets FLICKER25K, CIFAR-10 and NUS-WIDE respectively. We also find that the larger the bit length, the greater the improvement of DSG over other methods, which implies that DSG can generate more independent hash bits.

\begin{figure}
	\centering
	\subfigure[FLICKR25K]{
	\begin{minipage}{4cm}
	\centering
	\includegraphics[width=4cm]{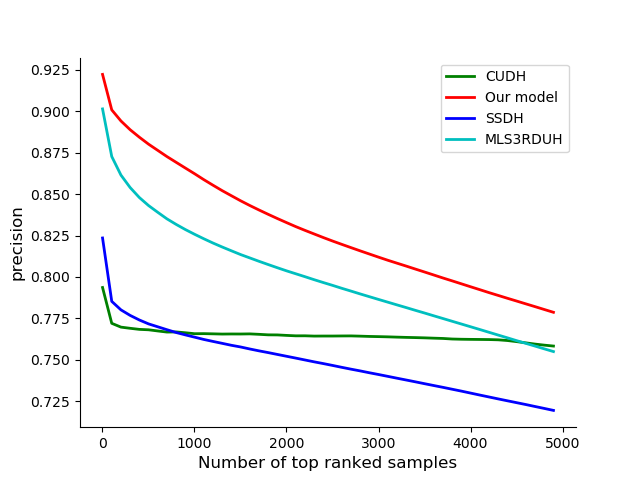}
	\end{minipage}
	}
	\subfigure[CIFAR-10]{
	\begin{minipage}{4cm}
	\centering
		\includegraphics[width=4cm]{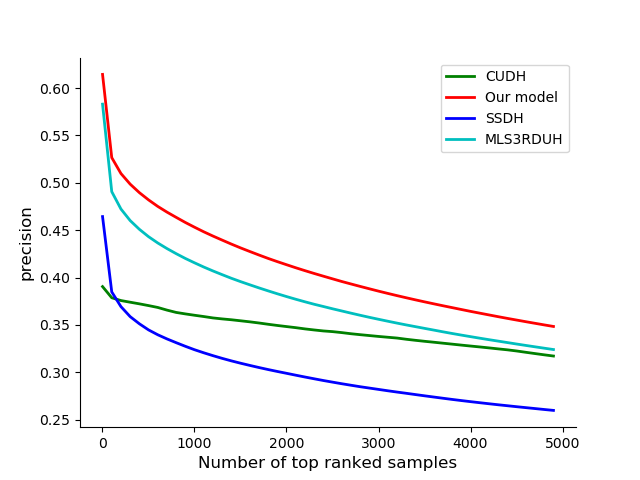}
	\end{minipage}
	}
	\subfigure[NUS-WIDE]{
	\begin{minipage}{4cm}
	\centering
		\includegraphics[width=4cm]{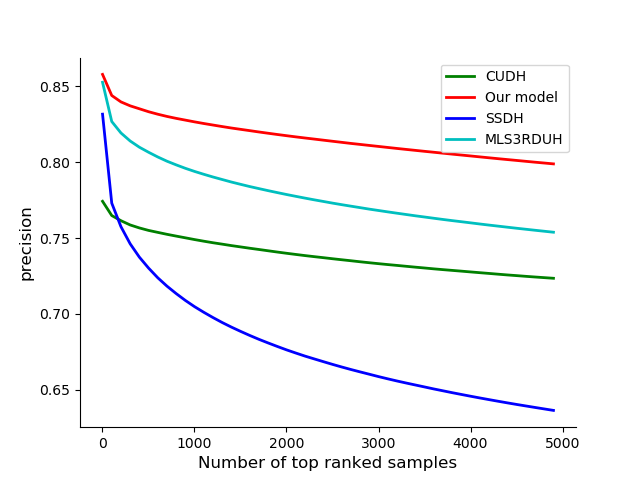}
	\end{minipage}
	}
	\subfigure[FLICKR25K]{
	\begin{minipage}{4cm}
	\centering
		\includegraphics[width=4cm]{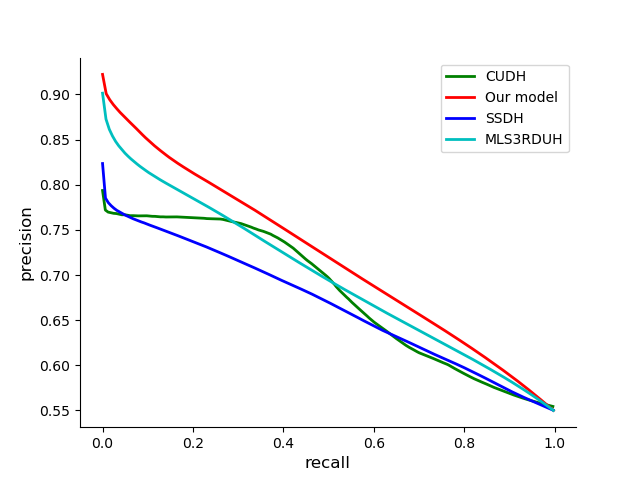}
	\end{minipage}
	}
	\subfigure[CIFAR-10]{
	\begin{minipage}{4cm}
	\centering
		\includegraphics[width=4cm]{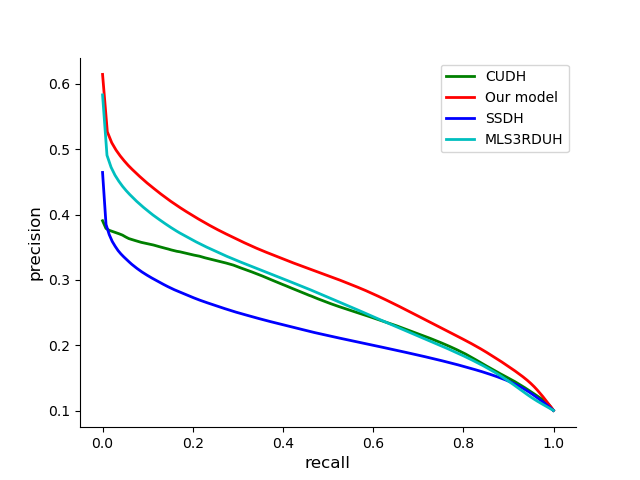}
	\end{minipage}
	}
	\subfigure[NUS-WIDE]{
	\begin{minipage}{4cm}
	\centering
		\includegraphics[width=4cm]{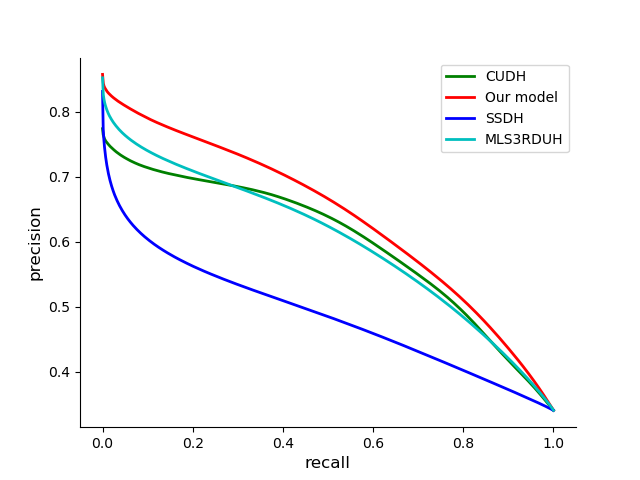}
	\end{minipage}
	}
\caption{(a), (b) and (c) are the Top N precision curves with code length 128 on FLICKR25K, CIFAR-10 and NUS-WIDE. (d) and (e) and (f) are the precision-recall curves with code length 128 on FLICKR25K, CIFAR-10 and NUS-WIDE.\label{fig:prc}}
\end{figure}

%

We also compare the performance of the DSG full model, the DSG model without the weight matrix $W_1$ , which is denoted as DSG-v1 and the DSG model without the weight matrix $W_1$ and the distilling matrix $W_2$, which is denoted as DSG-v2.
The results are shown in Table \ref{tab:2}. It is easy to find that DSG always achieves the highest MAPs, which implies that both the confidence information provided by the weight matrix $W_1$ and the global consistency that is ensured by the distilling matrix $W_2$ are necessary in our model. We can also find that DSG-v1 always outperforms DSG-v2, which proves that the clustering alone can help improve performance by removing contradictory similarity relationships between the pseudo-graph $S$ and the relationship matrix $C$.

For a more comprehensive comparison, we draw precision-recall curves and Top N precision curves for our method DSG and three state-of-art methods CUDH, SSDH and MLS$^3$RUDH with the hash code length of 128. Figure \ref{fig:prc} (a), (b) and (c) show the Top N precision curves of CUDH, DSG, SSDH and MLS$^3$RUDH on datasets FLICKR25K, CIFAR-10 and NUS-WIDE. It can be seen that DSG always has the highest precision among these four models and MLS$^3$RUDH always has the second-highest precision. Since the precision curves are based on the ranks of Hamming distance, DSG is able to achieve the highest recall if we directly use Hamming distance for retrieval. It's known that hash codes can also be used for coarse filtering in the form of hash table lookup; we also plot the precision-recall curves for these four models on the same datasets, which are shown in Figure \ref{fig:prc} (d), (e) and (f). It can be clearly seen that the curves of DSG are always on top of the other three models' curves, which implies that the hash codes obtained by DSG are also more suitable for the hash table lookup search strategy, which further demonstrates the superiority of our method. 


\begin{figure}[h]
\centering
\includegraphics[width=3.6in]{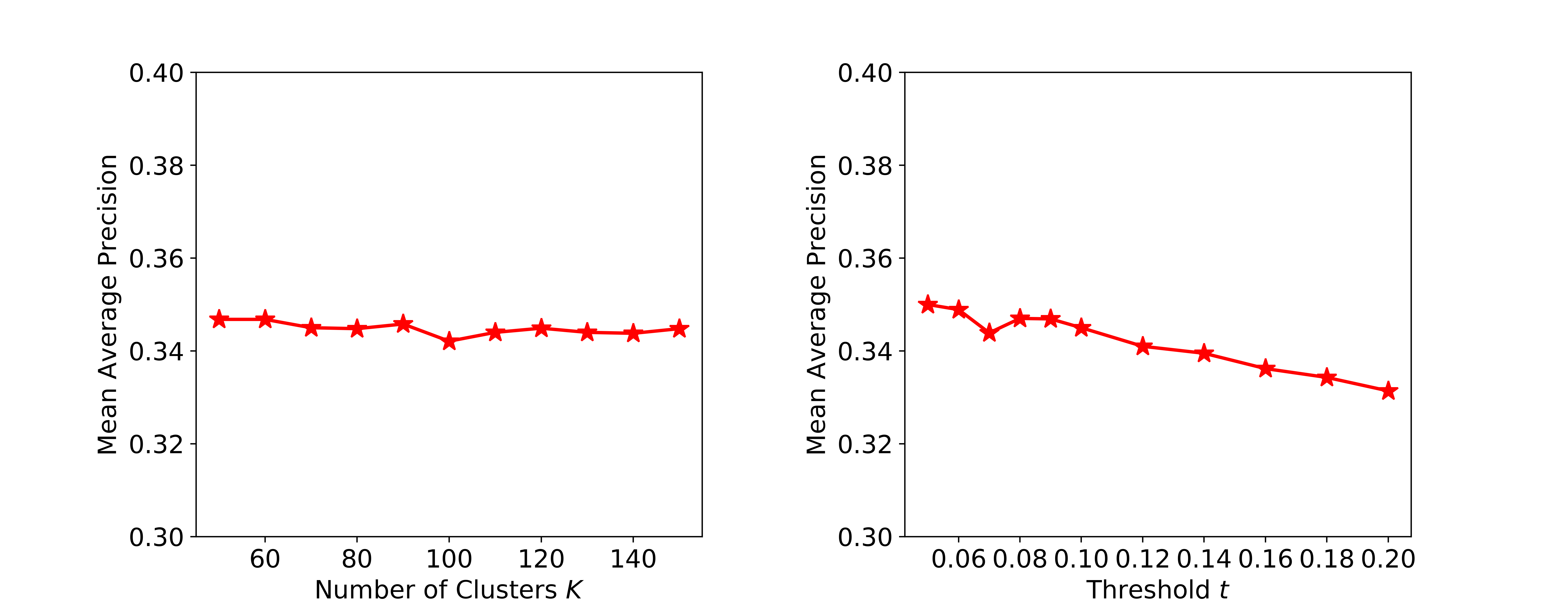}
\caption{MAP w.r.t different numbers of clusters and threshold $t$ values with code length 128 on CIFAR-10.\label{fig:sen2}}
\end{figure}

\subsection{Parameter Sensitivity}

We further study the number of clusters $K$ and threshold $t$. Figure \ref{fig:sen2} shows the MAPs of different $K$ values ranging $[50,150]$ on the dataset CIFAR-10 with code length 128 and the MAPs of different threshold $t$ values ranging $[0.05,0.20]$ on the dataset CIFAR-10 with code length 128 as well. We find the MAP with the $K$ value of 100 is slightly lower than other $K$ values'. In general, DSG's performance with different $K$ values ranging $[50,150]$ is relatively stable, indicating that the model performance is not sensitive to different $K$ values ranging $[50,150]$. Furthermore, we show that the MAP decreases as the threshold $t$ is over $0.1$ with K fixed to $70$.
In addition, the performance of DSG with different $t$ ranging $[0.05, 0.1]$ is relatively stable, which indicates that the suitable interval for $t$ value is $[0.05, 0.1]$. 
Accordingly, we set $K$ as $70$ and $t$ as $0.1$ in our other experiments as default.

\section{Conclusion}
In this paper, we proposed Distilled Smooth Guidance (DSG) for deep unsupervised hashing. DSG not only considers local similarity signals but also considers the confidence of similarity signals from local structure for smoothness. What's more, global information is also explored and the image pairs are distilled by removing contradictory image pairs from two views for the purpose of accuracy. Numeric experiments demonstrates that DSG outperforms the existing state-of-the-art methods.

\bibliographystyle{IEEEtran}
\bibliography{ms}

\end{document}